# Research of an optimization model for servicing a network of ATMs and information payment terminals

Nigmatulin G.A[1][0000-0001-8935-1146] and Chaganova O.B [2][0000-0001-9899-2527]

[1] Moscow Institute of Physics and Technology, Moscow, Russia
[2] Orenburg State University, Orenburg, Russia

**Abstract.** The steadily high demand for cash contributes to the expansion of the network of Bank payment terminals. To optimize the amount of cash in payment terminals, it is necessary to minimize the cost of servicing them and ensure that there are no excess funds in the network. The purpose of this work is to create a cash management system in the network of payment terminals. The article discusses the solution to the problem of determining the optimal amount of funds to be loaded into the terminals, and the effective frequency of collection, which allows to get additional income by investing the released funds. The paper presents the results of predicting daily cash withdrawals at ATMs using a triple exponential smoothing model, a recurrent neural network with long short-term memory, and a model of singular spectrum analysis. These forecasting models allowed us to obtain a sufficient level of correct forecasts with good accuracy and completeness. The results of forecasting cash withdrawals were used to build a discrete optimal control model, which was used to develop an optimal schedule for adding funds to the payment terminal. It is proved that the efficiency and reliability of the proposed model is higher than that of the classical Baumol-Tobin inventory management model: when tested on the time series of three ATMs, the discrete optimal control model did not allow exhaustion of funds and allowed to earn on average 30% more than the classical model.

## 1       Introduction

Despite the rapid development of Internet banking, the demand for cash has remained high for many years. In this regard, the size of the network of payment terminals of banks is also increasing. Some financial institutions may have thousands of terminals that are expensive to maintain: controlling the network and making money management decisions entails paying employees and significant operating costs (financial, transportation, loading and unloading, insurance, etc.). It is equally important to properly manage the cash available in the terminals – to ensure that there are no excess funds circulating in the network.

When optimizing the amount of cash in the terminals, the main problem is the unpredictability of the daily withdrawal level. The demand for cash for a particular ATM depends on many factors: salary days, holidays, seasonal demand, location specifics, and others.

Thus, banks have two tasks. The first task is to determine the exact amount of funds that need to be loaded into the terminals in order to provide customers with the required amount of cash. To solve this problem, we consider models for predicting daily cash withdrawals at ATMs. The second task is to determine, based on the obtained forecast, the effective collection frequency that maximizes the difference between the investment income of the released funds from the "cash deposits" of the terminal and the cost of its maintenance. The discrete optimal control model is used as an approach for solving this problem.

## 2   Forecasting time series of daily cash withdrawals

### 2.1   Data and quality metrics

Data for the study is taken from the open source Kaggle. They represent the history of cash withdrawals from January 01, 2011 to June 11, 2017 at three ATMs in Bangalore, India. The initial sample was divided into training and test samples: 2340 days were included in the training sample, and the remaining 14 days were included in the test sample. To demonstrate the application of forecasting methods, the time series of the "Christ College ATM" terminal was selected.

To assess the quality of the forecast is calculated the error matrix, which consists of the following values: True Positive (TP) – a predicted amount of cash withdrawals is corresponded to reality; False Positive (FP) – error of the first type, the projected amount of cash has been overrated; False Negative (FN) – error of the second type, the forecast was lower than the demanded volume of cash; True Negative (TN) – the projected cash balance in the terminal is coincided with reality. When predicting cash withdrawals, a second type of error plays an important role: if there is a shortage of cash reserves inside the terminal, the client will not receive the required amount of cash, which may negatively affect their loyalty to the Bank.

### 2.2   Forecasting using the triple exponential smoothing model

In the case of forecasting the time series of cash withdrawals in payment information terminals, which contains a pronounced trend and seasonal components, a triple exponential smoothing model can be applied [1]. It is assumed that the trend has an additive character, while the seasonal component has a multiplicative character with a periodicity equal to two weeks, i.e. the length of the season is 14 days.

As a result of the forecast, the estimated cash withdrawals values for the next 14 days from the forecast date were obtained, which are shown in figure 1. to assess the accuracy of the constructed forecast values, an error matrix was constructed (table 1) and the main forecast metrics were calculated (table 2).

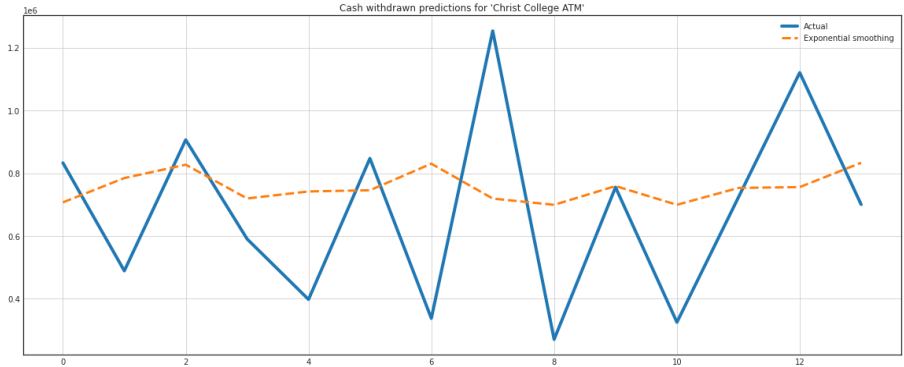

**Fig. 1.** The true and predicted with triple exponential smoothing values of cash withdrawals.

**Table 1.** Confusion matrix for triple exponential smoothing method.

|  |  | Actual values | |
|---|---|---|---|
|  |  | All money was withdrawn | Money left in the terminal |
| Predictable values | All money was withdrawn | 6 820 481 | 2 230 881 |
|  | Money left in the terminal | 1 204 478 | 3 755 122 |

**Table 2.** Metrics of the triple exponential smoothing model.

| Metric | Accuracy | Precision | Recall | F1 |
|---|---|---|---|---|
| Value | 0,8517 | 0,8571 | 1,000 | 0,9231 |

### 2.3 Prediction using recurrent neural networks with long short-term memory

Data on cash withdrawals in ATMs and information payment terminals is a time sequence, which is predicted using recurrent neural networks with long short-term memory. To do this, train the neural network to predict the expected level of withdrawals one day ahead based on the previous historical data of one month, i.e. 31 days. This operation can be recursively applied for forecasting for a longer period of time, assuming that the values of previous forecasts will be used at each new forecasting step [2]. During the experiments, the highest accuracy was provided by a neural network model with two hidden layers consisting of 50 neurons. The rectified linear activation function (ReLU) is selected as the activation function for each neuron on each layer.

To avoid possible retraining and regularization of the neural network, we used Dropout thinning layers for each hidden layer, which in 20% of cases "discarded" a number of output properties of each layer randomly during training. During the training, the adam optimization function was used, and the mean square error MSE was used as the loss function. The number of epochs is 10.

As a result of the forecast, the estimated values of cash withdrawals over the next 14 days from the forecast date were obtained, which are shown in figure 2. to assess the accuracy of the constructed forecast values, an error matrix was constructed (table 3) and the main forecast metrics were calculated (table 4).

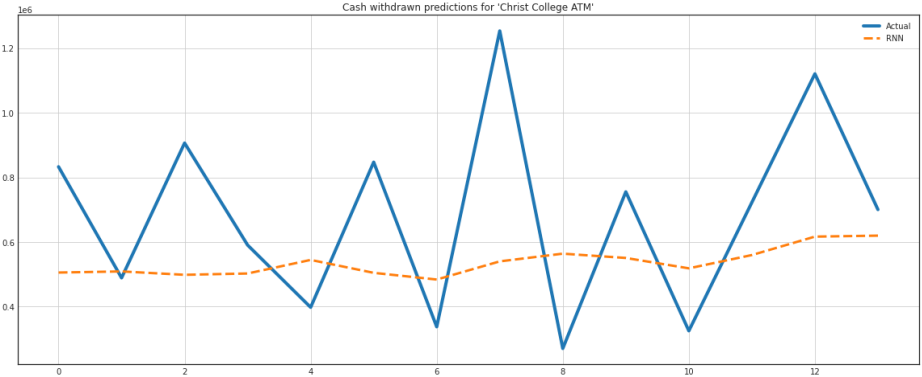

**Fig. 2.** The true and predicted with a recurrent neural network with long short-term memory values of withdrawals of cash.

**Table 3.** Confusion matrix for recurrent neural networks with long short-term memory.

|  |  | Actual values | |
|---|---|---|---|
|  |  | All money was withdrawn | Money left in the terminal |
| Predictable values | All money was withdrawn | 2 620 473 | 799 673 |
|  | Money left in the terminal | 2 829 494 | 4 898 906 |

**Table 4.** Metrics of the recurrent neural networks with long short-term memory model.

| Metric | Accuracy | Precision | Recall | F1 |
|---|---|---|---|---|
| Value | 0,8571 | 1,000 | 0,833 | 0,9091 |

### 2.4 Prediction using singular spectral analysis

The problem of predicting cash withdrawals in a payment terminal is difficult mainly because it is impossible to make an unambiguously correct assumption about the frequency that is inherent in this process. Therefore, using the method of singular spectral analysis, which initially does not make assumptions about periodicity, can be extremely effective [3].

The window length is equal to half the size of the training sample, i.e. 1177. As a result of constructing the delay matrix and singular value decomposition, 568 eigenvectors were obtained, where their eigenvalues are considered as contributions of the main ones to the original time series (figure 3).

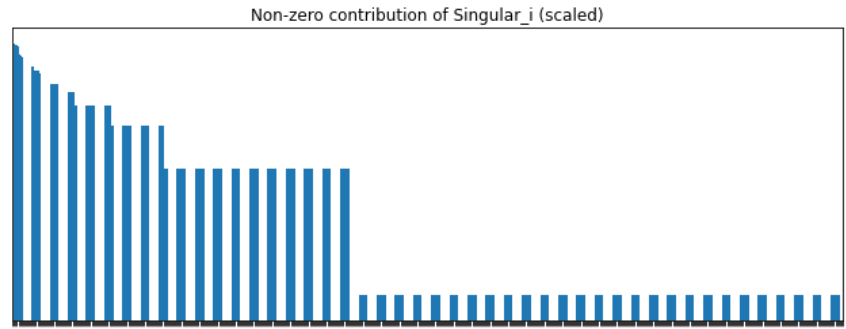

**Fig. 3.** Eigenvalues of the trajectory matrix of the original time series

To restore the delay matrix and the original time series, 500 eigenvectors were selected out of 568. As a result, a reconstructed series is obtained that repeats the main regularities of the original series with high accuracy (figure 4).

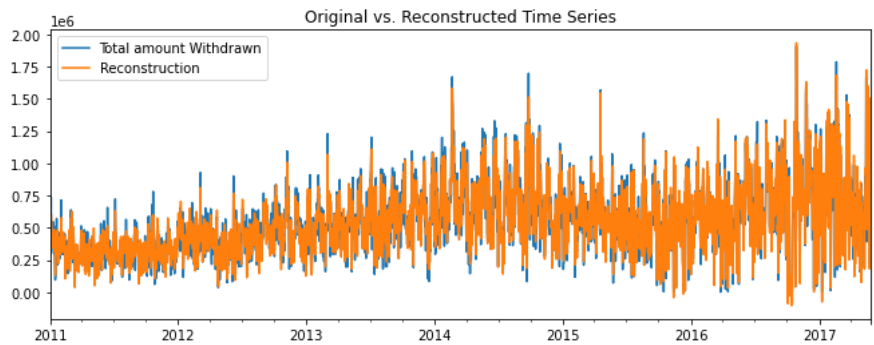

**Fig. 4.** The true and predicted with the singular spectrum analysis time series.

The method of singular spectral decomposition involves predicting one value ahead. Therefore, a recurrent approach is used, where each obtained forecast value is used as input data for predicting the next value. As a result of the forecast, the estimated values of withdrawals over the next 14 days from the forecast date were obtained, which are shown in figure 5. To assess the accuracy of the forecast values, an error matrix was constructed (table 5) and the main forecast metrics were calculated (table 6).

**Table 5.** Confusion matrix for singular spectral analysis.

|  |  | Actual values | |
| --- | --- | --- | --- |
|  |  | All money was withdrawn | Money left in the terminal |
| Predictable values | All money was withdrawn | 2 620 473 | 799 673 |
|  | Money left in the terminal | 2 829 494 | 4 898 906 |

**Table 6.** Metrics of the recurrent neural networks with long short-term memory model.

| Metric | Accuracy | Precision | Recall | F1 |
|---|---|---|---|---|
| Value | 0,8571 | 1,000 | 0,833 | 0,9091 |

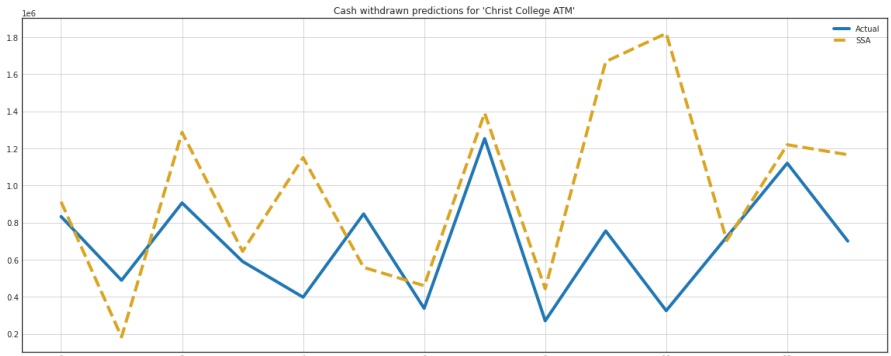

**Fig. 5.** The true and predicted with singular spectrum analysis values of withdrawals of cash.

## 2.5 Comparative analysis of the forecasting results

Forecasting the time series of cash withdrawals in the terminal for 14 days ahead was performed using three models: the triple exponential smoothing model (EXP), the neural network model with long short-term memory (LSTM), and the singular spectral analysis model (SSA). Let's compare the results to choose the model that is best suited to the task at hand.

The results of forecasting using the considered models are shown in figure 6. the model is selected using a table with aggregated values of error matrices (table 7) and a table with aggregated metrics of forecasts using various methods (table 8).

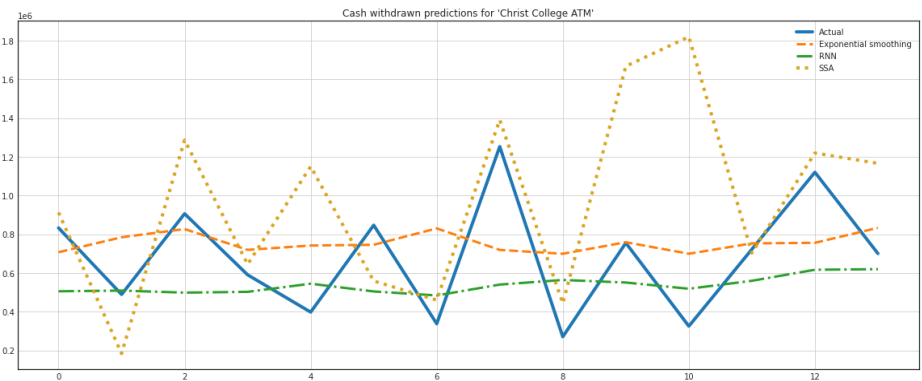

**Fig. 6.** Results of forecasting using ESP, LTD and SSA models.

**Table 7.** Metrics of the recurrent neural networks with long short-term memory model.

|  |  | Actual values ||||||
|---|---|---|---|---|---|---|---|
|  |  | EXP || LSTM || SSA ||
|  |  | All money was withdrawn | Money left in the terminal | All money was withdrawn | Money left in the terminal | All money was withdrawn | Money left in the terminal |
| Values | All money was withdrawn | 6820481 | 2230881 | 2620473 | 799673 | 2620473 | 799673 |
| Values | Money left in the terminal | 1204478 | 3755122 | 2829494 | 4898906 | 2829494 | 4898906 |

**Table 8.** Metrics of the recurrent neural networks with long short-term memory model.

|  | EXP | LSTM | SSA |
|---|---|---|---|
| Accuracy | 0,8517 | 0,8571 | 0,7143 |
| Precision | 0,8571 | 1,000 | 0,7667 |
| Recall | 1,000 | 0,833 | 1,0000 |
| F1 | 0,9231 | 0,9091 | 0,8679 |

All the methods considered allow to get a sufficient level of correct forecasts with good accuracy and completeness. However, in the case of forecasting cash withdrawals, the second type of error plays a big role, so the Accuracy and Recall metrics are more important when making a decision about choosing a forecasting method. Therefore, predicting cash withdrawals in the terminal using the singular spectral analysis method is better than other methods considered to avoid unexpected exhaustion of cash reserves inside the terminal.

## 3 Building an optimal schedule for adding funds to payment terminals

### 3.1 The system of the discrete optimal control

With forecasted cash withdrawals for 2 weeks ahead, the second part of the system under study becomes possible – managing the schedule and replenishment volumes of terminals, which will be implemented by a discrete optimal control system:

$$\begin{cases} W(T, x(T), y(T)) \to \max & (1) \\ W(t, x(T), y(T)) = W^0 + \sum_{i=0}^{t-1}((1 + RuOnIa) \cdot W(i) - (x(i) + P \cdot y(i))) & (2) \\ s(t) = s^0 + \sum_{i=0}^{t-1}(x(i) - l(i)) & (3) \\ s(t) \geq \alpha \cdot d(t) & (4) \\ s(t) + x(t) \leq C & (5) \\ W^0 = \sum_{i=0}^{T} l(i) & (6) \\ P, RuOnIa, s^0, \alpha, C - \text{const} & (7) \end{cases}$$

where $t = \overline{1, T}$ – sequence of days for optimization; $W(t)$ – balance of funds in the Bank for replenishment of terminals; $W^0$ – initial level of managed funds in the bank; $P$ – cost of collection; $y(t)$ – act of collection on the t-th day; RuOnIa – rate of return on one – day ruble deposits «overnight»; $x(t)$ – amount of replenishment on the t – th day; $s(t)$ – amount of funds in the terminal; $s^0$ – initial amount of funds in the terminal; $l(t)$ – expected withdrawals for the t – th day; $\alpha$ – security coefficient of the lower limit of replenishment; $d(t)$ – standard deviation of expected cash withdrawals on the t-th day; $C$ – terminal capacity.

The solution of the system will be the optimal trajectory of daily replenishment volumes of the payment terminal to achieve the maximum balance of funds in the Bank W at the final time T, with restrictions on the set of acceptable values of replenishment volumes (3-7). The initial conditions are shown in table 9.

**Table 9.** Initial conditions.

| Indicator | Decryption | Value |
|---|---|---|
| P | Cost of collection | 1000 rupee |
| RuOnIa | Rate of return | 0.0001567 |
| $s^0$ | Initial level of funds in the terminal | 0 rupee |
| C | The capacity of the terminal | 13 000 000 rupee |
| α | Safety factor | 0.2 |

### 3.2 The system of the discrete optimal control

Before making an optimal plan, it is necessary to evaluate the effectiveness of each collection. To do this, a matrix of cash requirements on day i for j days ahead is constructed from the forecast values (figure 7).

|   | 1 | 2 | 3 | 4 | 5 |
|---|---|---|---|---|---|
| 0 | [912522.22107146] | [1096341.91341871] | [2382928.6150902603] | [3027816.0200476204] | [4178011.4746250105] |
| 1 | 0 | [183819.69234725] | [1470406.3940188] | [2115293.79897616] | [3265489.25355355] |
| 2 | 0 | 0 | [1286586.70167155] | [1931474.1066289102] | [3081669.5612063003] |
| 3 | 0 | 0 | 0 | [644887.40495736] | [1795082.8595347502] |
| 4 | 0 | 0 | 0 | 0 | [1150195.45457739] |
| 5 | 0 | 0 | 0 | 0 | 0 |

**Fig.7.** Matrix of cash requirements on day i for j days ahead.

To determine the maximum revenue collection plan, we will use the principle of dynamic programming. In this way, a matrix of capital values can be calculated when depositing funds on day i for j days in advance and investing funds held in the Bank (figure 8), and a matrix of income from investing funds held in the Bank when depositing funds on day i for j days in advance (figure 9).

|   | 1 | 2 | 3 | 4 | 5 |
|---|---|---|---|---|---|
| 0 | [12697265.27451135] | [12515377.782035738] | [11230147.427571421] | [10586615.582906026] | [9437177.649951333] |
| 1 | 0 | [12517367.911874894] | [11232310.736128192] | [10588880.339807283] | [9439623.124592172] |
| 2 | 0 | 0 | [11234301.489771986] | [10590871.405426716] | [9441665.361802379] |
| 3 | 0 | 0 | 0 | [10592733.328360463] | [9443657.05151977] |
| 4 | 0 | 0 | 0 | 0 | [9445497.629879382] |
| 5 | 0 | 0 | 0 | 0 | 0 |

**Fig. 8.** Matrix of capital values when adding funds on day i for j days in advance and investing funds held in the bank.

|   | 1 | 2 | 3 | 4 | 5 | 6 | 7 |
|---|---|---|---|---|---|---|---|
| 0 | 989.506 | 2921.71 | 4278.05 | 5633.61 | 6391.14 | 7344.46 | 8231.05 |
| 1 | 0 | 961.626 | 2520.28 | 3977.6 | 4916.15 | 5957.8 | 6917.27 |
| 2 | 0 | 0 | 760.554 | 2319.24 | 3438.42 | 4568.01 | 5599.97 |
| 3 | 0 | 0 | 0 | 660.012 | 1959.76 | 3177.24 | 4281.64 |
| 4 | 0 | 0 | 0 | 0 | 480.226 | 1785.54 | 2962.31 |

**Fig. 9.** Matrix of income from investing funds held in the Bank, when adding funds to the terminal on the i-th day for o days in advance.

Having a matrix of income from investing funds, the problem of finding the optimal strategy that ensures maximum profit can be represented as the problem of finding the shortest path in a graph whose weights are values from this matrix with minus signs. To solve this problem, use the Bellman-Ford algorithm. As a result of the algorithm, the sequence of days that must be collected is calculated (figure 10).

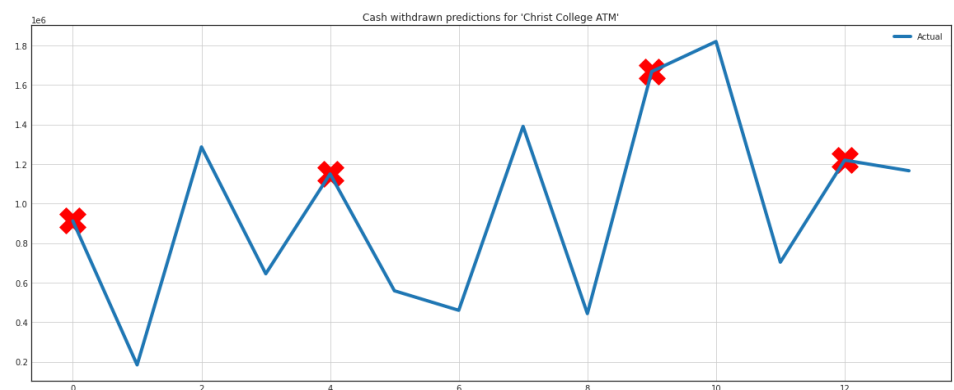

**Fig. 10.** The sequence of days when cash collection should be performed.

### 3.3 Comparative analysis of the results obtained by the method of discrete optimization and classical models of inventory management

To compare the effectiveness of the discrete optimal control model and the classical Baumol-Tobin inventory management model, an experiment was performed (figure 11). It was found that for a given security level, the proposed model does not allow terminal exhaustion, while the Baumol-Tobin model with an economical replenishment order size of 2616761.89 rupees and a lower limit of 740428.11 rupees allowed 4 exhaustions. Using the proposed model of discrete optimal control, we managed to earn 25.81% more than using the Baumol-Tobin model.

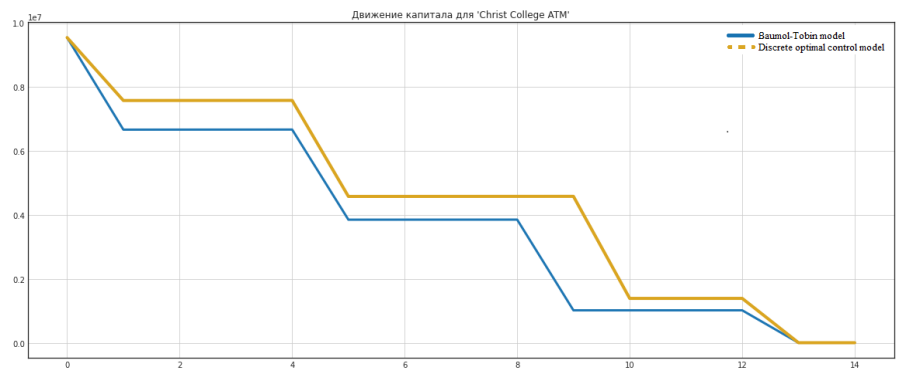

**Fig. 11.** Capital flows in cases of usage the Baumol-Tobin model and optimal control model.

To confirm the generality of the results obtained, the proposed model of forecasting and discrete optimal control was applied to the cash withdrawal series "KK Nagar ATM" and "Big Street ATM" (figure 12). The model demonstrated reliability without running out of funds, while the Baumol-Tobin model allowed 3 and 1 exhaustion, respectively, and high efficiency, earning 26% and 36% more, respectively.

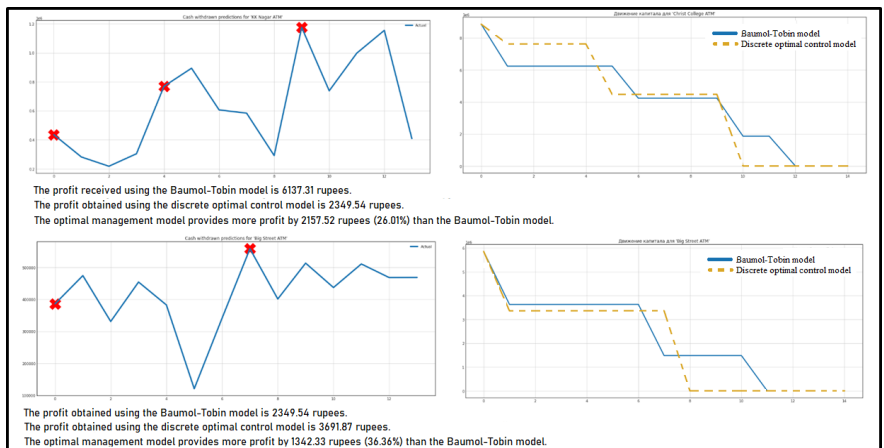

**Fig. 12.** Capital flows in cases where the Baumol-Tobin model and the discrete optimal control model are applied (cash withdrawals series "KK Nagar ATM" and " Big Street ATM»).

## 4      Conclusion

In this study, a cash collection management system was implemented that provides for predicting the time series of daily cash withdrawals in payment terminals and using this forecast to optimize the schedule of replenishment of terminals using a discrete optimal control model. The optimal solution to the problem of managing cash reserves in the network of payment terminals is a schedule (the number of days on which it is necessary to carry out collection), and the amount of cash that provides a sufficient level of funds in the terminal and a high level of income from investing funds remaining in the Bank.

   Computational experiments have shown that all the forecasting models described in this paper allow us to obtain a sufficient level of correct forecasts with good accuracy and precision. However, the quality of the forecast models depends on the source data – each terminal has an individual nature of cash withdrawals, and therefore it is necessary to conduct separate studies for each of them. For example, the method of singular spectral analysis was most effective for the "Christ College ATM" and "Airport ATM" terminals, while the model of a recurrent neural network with long short – term memory was used for the "KK Nagar ATM" terminal.

   When comparing the proposed model of forecasting and discrete optimal control with the classical Baumol-Tobin inventory management model, the proposed model demonstrates higher reliability and efficiency indicators. As a result of the experiments, it was revealed that the proposed system allows you to manage the collection process more reliably and efficiently, earning an average of 30% more money from investing in short-term financial instruments and completely eliminating the possibility of running out of funds in the terminal, compared to the classic cash management model. From an economic point of view, when using the considered model for managing cash reserves in a network of information payment terminals, for one terminal, the ATM in India considered, revenue per year will be about 200 thousand rupees, while for an urban network consisting, on average, of 50 terminals, this amount will be 11.5 million rupees.

## References

1. Rafi, M. [et al.]. ATM Cash Prediction Using Time Series Approach // 2020 3rd International Conference on Computing, Mathematics and Engineering Technologies (iCoMET), Sukkur, Pakistan. 2020. P. 1-6.
2. Kamini, V. [et al.]. Cash demand forecasting in ATMs by clustering and neural networks // European Journal of Operational Research. Vol. 232. Issue 2. 2014. P. 383-392.
3. Alvarez-Mesa, A.M. [et al.]. Automatic Singular Spectrum Analysis for TimeSeries Decomposition // European Symposium on Artificial Neural Networks, Computational Intelligence and Machine Learning. 2013. P. 131-136.
4. Catal, C. Improvement of demand forecasting models with special days / C. Catal, A. Fenerci, B. Ozdemir, O. Gulmez, // Proc. Comput. Sci. – 2015. - №59. – Pp. 262–267.
5. Aseev, M. [et al.]. Forecasting cash withdrawals in the ATM network using a combined model based on the holt-winters and markov chains // International Journal of Applied Engineering Research. 11. 2016. P. 7577–7582.

6. Simutis, R. [et al.]. Cash Demand Forecasting for ATM using Neural Networks and Support Vector Regression Algorithms // 20th EURO Mini Conference "Continuous Optimization and Knowledge-Based Technologies". 2008. P. 416-421.
7. Borda, P. [et al.]. Optimization of ATM filling-in with cash // 99th European study group with industry. 2014. P. 1–16.
8. Golyandina, N. Analysis of Time Series Structure: SSA and Related Techniques / N. Golyandina, V. Nekrutkin, A. A Zhigljavsky // CRC Press - 2001. – Pp. 17-44.
9. Venkatesh, K. Cash Demand Forecasting in ATMs by Clustering and Neural Networks / K. Venkatesh, V. Ravi, A. Prinzie, D. Van den Poel // European Journal of Operational Reseach. – 2013. – Pp. 383-392.
10. Acuna, G. Comparing NARX and NARMAX models using ANN and SVM for cash demand forecasting for ATM / G. Acuna, C. Ramirez, M. Curilem // IEEE World Congress on Computational Intelligence. - 2012. - Pp. 1-6.
11. Baumol, W. J. The Transactions Demand for Cash: An Inventory Theoretic Approach / W. J. Baumol // Quarterly Journal of Economics. - 1952. -№66. – Pp. 545-556.
12. Snellman, H. ATM networks and cash usage / H. Snellman, M. Viren // Research Discussion Papers. – 2006. - №.21. – Pp. 1-33.

About the authors:

**Georgii A. Nigmatulin**, master of the Department of information transfer and data analysis, Moscow Institute of physics and mathematics (9, Institutsky per., Dolgoprudny, 141701, Russia), ORCID: http://orcid.org/0000-0001-8935-1146, george_nigm@icloud.com

**Olga B. Chaganova**, student of the Department of mathematical methods and models in Economics, Orenburg state University (13 Pobedy Ave., Orenburg, 460018, Russia), ORCID: https://orcid.org/0000-0001-9899-2527, olga.chaganova1@gmail.com